\documentclass[sigconf,authorversion,nonacm]{acmart}

\usepackage{hyperref}

\usepackage{subcaption}
\usepackage{multirow}
\usepackage{booktabs}
\AtBeginDocument{%
  \providecommand\BibTeX{{%
    \normalfont B\kern-0.5em{\scshape i\kern-0.25em b}\kern-0.8em\TeX}}}

\begin{document}

\title[Confidence-Aware Sub-Structure Beam Search (CABS): Mitigating Hallucination in Structured Data Generation with Large Language Models] {Confidence-Aware Sub-Structure Beam Search (CABS): Mitigating Hallucination in Structured Data Generation with Large Language Models}

\author{Chengwei Wei}
\authornote{This work is done during an internship at Amazon}
\email{chengwei@usc.edu}
\affiliation{%
  \institution{University of Southern California}
  \country{USA}
}

\author{Kee Kiat Koo}
\email{kiatkoo@amazon.com}
\affiliation{%
  \institution{Amazon}
  \country{USA}}

\author{Amir Tavanaei}
\email{atavanae@amazon.com}
\affiliation{%
  \institution{Amazon}
  \country{USA}}

\author{Karim Bouyarmane}
\email{bouykari@amazon.com}
\affiliation{%
  \institution{Amazon}
  \country{USA}
}

\begin{teaserfigure}
\begin{center}
\huge
\textcolor{blue}{\url{https://cabsllm.github.io}}
\end{center}
\vskip 1.5em
\end{teaserfigure}

\begin{abstract}

Large Language Models (LLMs) have facilitated structured data generation, with applications in domains like tabular data, document databases, product catalogs, etc. However, concerns persist about generation veracity due to incorrect references or hallucinations, necessitating the incorporation of some form of model confidence for mitigation. Existing confidence estimation methods on LLM generations primarily focus on the confidence at the individual token level or the entire output sequence level, limiting their applicability to structured data generation, which consists of an intricate mix of both independent and correlated entries at the sub-structure level. In this paper, we first investigate confidence estimation methods for generated sub-structure-level data. We introduce the concept of Confidence Network that applies on the hidden state of the LLM transformer, as a more targeted estimate than the traditional token conditional probability. We further propose Confidence-Aware sub-structure Beam Search (CABS), a novel decoding method operating at the sub-structure level in structured data generation. CABS enhances the faithfulness of structured data generation by considering confidence scores from the Confidence Network for each sub-structure-level data and iteratively refining the prompts. Results show that CABS outperforms traditional token-level beam search for structured data generation by 16.7\% Recall at 90\% precision averagely on the problem of product attribute generation.

\end{abstract}

%%
%% The code below is generated by the tool at http://dl.acm.org/ccs.cfm.
%% Please copy and paste the code instead of the example below.
%%

% \begin{CCSXML}
% <ccs2012>
%    <concept>
%        <concept_id>10002944.10011122.10002947</concept_id>
%        <concept_desc>General and reference~General conference proceedings</concept_desc>
%        <concept_significance>500</concept_significance>
%        </concept>
%    <concept>
%        <concept_id>10010147.10010178.10010179</concept_id>
%        <concept_desc>Computing methodologies~Natural language processing</concept_desc>
%        <concept_significance>500</concept_significance>
%        </concept>
%    <concept>
%        <concept_id>10010147.10010257</concept_id>
%        <concept_desc>Computing methodologies~Machine learning</concept_desc>
%        <concept_significance>500</concept_significance>
%        </concept>
%    <concept>
%        <concept_id>10010405.10010481.10003558</concept_id>
%        <concept_desc>Applied computing~Consumer products</concept_desc>
%        <concept_significance>300</concept_significance>
%        </concept>
%  </ccs2012>
% \end{CCSXML}

% \ccsdesc[500]{General and reference~General conference proceedings}
% \ccsdesc[500]{Computing methodologies~Natural language processing}
% \ccsdesc[500]{Computing methodologies~Machine learning}
% \ccsdesc[300]{Applied computing~Consumer products}

%%
%% Keywords. The author(s) should pick words that accurately describe
%% the work being presented. Separate the keywords with commas.

%%
%% This command processes the author and affiliation and title
%% information and builds the first part of the formatted document.
\maketitle

\section{Introduction}

The development of Large Language Models (LLMs) has made significant advances over the last few years, especially in text generation. Several recent studies have explored the ability of LLMs to output structured data such as product catalogs~\cite{roy2021attribute, roy2022exploring, blume2023generative}, tabular data~\cite{borisov2022language, wu2022text, li2023sequence, kitouni2023kbformer}, and programming languages~\cite{wang2021codet5, ahmad2021unified, zhang2023planning} as opposed to natural-language text generation.
Unlike plain text, structured data generation conforms to a standardized output format according to some schemas. For example, in the medical domain, although the raw input data is textual in the form of clinical notes, the underlying structure is tabular in nature, consisting of a patient's demographics and other medical diagnostic attributes.

\begin{figure*}[ht]
    \includegraphics[width=14.5cm]{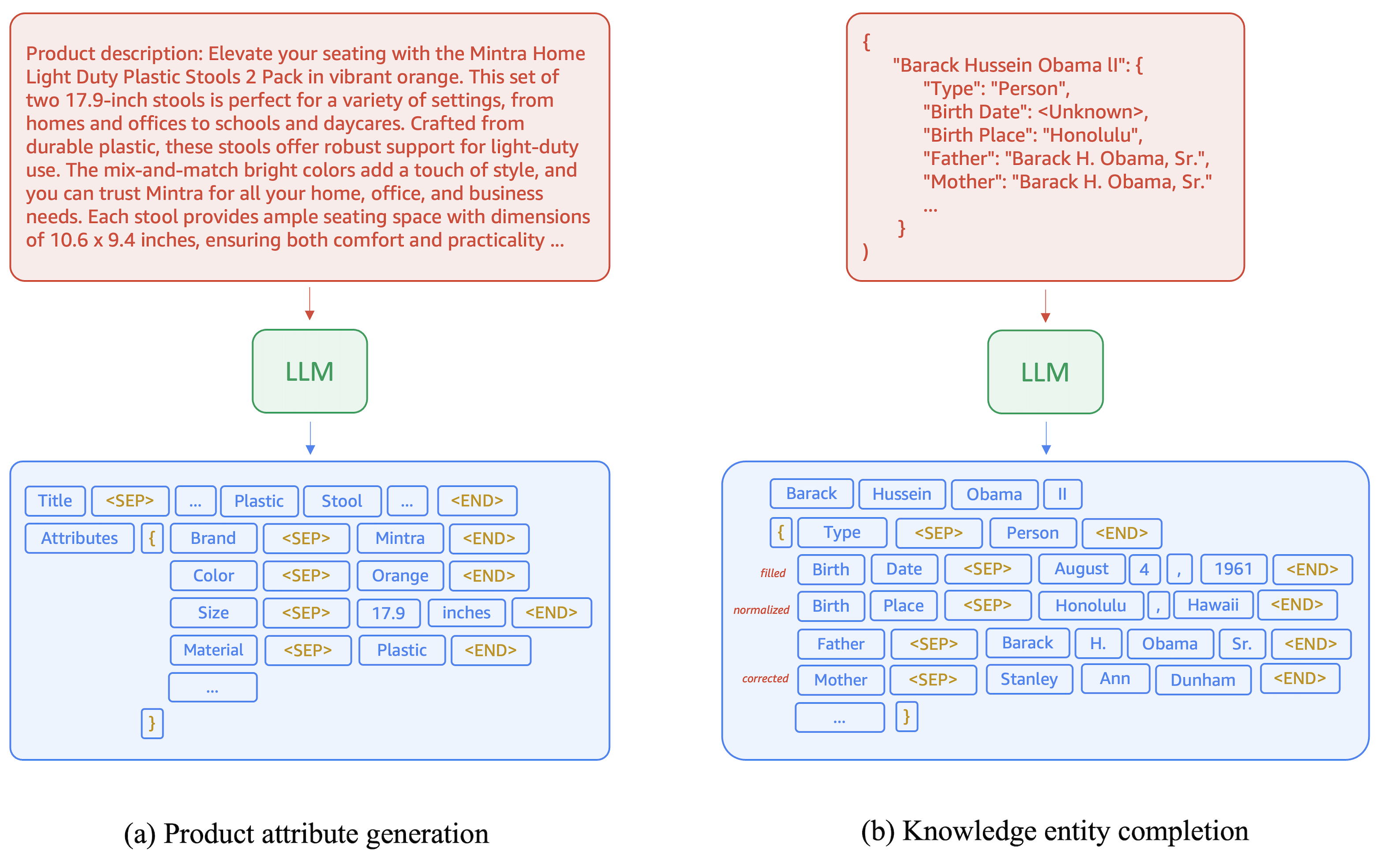}
    \caption{Examples of structured data generation using LLMs.}
    \label{fig:example_structured_data_generation}
\end{figure*}

Structured data generation can be processed using a single prompt. This is typically achieved by fine-tuning an LLM specifically into a Knowledge base \cite{kitouni2023kbformer}. Instead of employing multi-pass generation, where each sub-structure-level data is processed individually, a Knowledge Base LLM jointly generates the entire structured data containing all sub-structures in a single prompt.
As an example in Fig. \ref{fig:example_structured_data_generation} (a), products in a catalog typically contain different attributes (i.e., sub-structures), including brand, color and size. The generation of each product attribute can be seen as a standalone sub-structure-level task. Although it is possible to prompt the LLM to generate each attribute individually via multiple passes, a Knowledge Base LLM regenerates all attributes as a whole, leading to a set of complete and detailed product attributes. Another example is knowledge entity completion where each entity can be regarded as a sub-structure. LLMs are prompted with a low-quality knowledge base to correct and complete all the knowledge entities.

%%%%%%%%%%%%%%%%%%%%%%%%%%%%%%%%%%%%%%%%%%%%%%%%%%%%%%%%%%%%%%%%%%
%%%%%%%%%%%%%%%%%%%%%%%%%%%%%%%%%%%%%%%%%%%%%%%%%%%%%%%%%%%%%%%%%%

Despite the achievements of LLMs, concerns persist regarding generation veracity, as LLMs may often make unfaithful generation due to hallucinations \cite{tian2019sticking,xiao2021hallucination,kadavath2022language, azaria2023internal, li2023helma}. This challenge extends beyond general unstructured data generation to structured data generation, necessitating the estimation of model confidence---the probability of the model's prediction being correct---in structured data generation to mitigate inaccuracies stemming from hallucinations. 
However, existing literature on confidence estimation methods primarily focuses on complete sequence-level~\cite{malinin2020uncertainty, jiang2021can, kadavath2022language} or individual token-level~\cite{xiao2021hallucination, zhou2021detecting, petryk2024simple}. These methods detect the confidence score of either the entire generated sequence or each generated token and are not the most appropriate method for structured data generation, where the LLM may be confident only of certain sub-structure generations. Retaining part of the correct sub-structure generation is preferable to discarding the entire generation. Thus, there is a need to differentiate within the LLM's response, identifying generated sub-structures in which the LLM is confident versus where it is not. 

In this paper, we use a Product Catalog LLM that is fine-tuned and able to generate all product attributes in a single pass to investigate confidence estimation methods on structured data generation. Specifically, the methods built on token conditional probability and the internal hidden state are examined. The experiments conducted on product catalog data from an online shopping website reveal that the internal states of LLMs are capable of assessing the faithfulness of the generated sub-structure. Building upon these insights, we introduce a novel decoding method for structured data generation named Confidence-Aware sub-structure Beam Search (CABS). CABS works on the sub-structure level in structured data generation, and examines the LLM confidence for each sub-structure and revises the prompt during generation to improve LLM output. In our experiments, we found that CABS significantly outperforms other traditional text generation decoding methods. 
Importantly, CABS is applicable to the generation of various structured data, beyond the structured data used in this paper. Our main contributions are summarized below.

\begin{itemize}
     \item We extend existing LLM confidence estimation methods to structured data generation, and introduce the Confidence Network as a new module in the LLM Transformer architecture.
     \item We propose CABS decoding, which relies on the confidence scores from the confidence estimation method when generating the next sub-structure-level data. 
\end{itemize}

The rest of the paper is organized as follows. In Section~\ref{sec: methods}, we study confidence estimation methods on generated sub-structure-level data and introduce the CABS approach. Section~\ref{sec: data and model} overview the data and model training details. Experimental results and analysis are presented in Section~\ref{sec: experiments}. Section~\ref{sec: related work} introduces related work on confidence estimation and text generation decoding. Finally, conclusions and future work are discussed in Section~\ref{sec: conclusion}.

\section{Method} \label{sec: methods}

\subsection{Preliminary}

In this work, we denote $x$ as the initial conditioning context (initial prompt) for a text generative LLM, and $y$ as the entire sequence of generated structured data. $t$ represent a token, while $t_{<i}$ indicates the tokens generated before token $t_i$.  Furthermore, $s$ denotes a sub-structure sequence, representing a sub-structure-level data, within $y$. Consider $y$ as a sequence of structured data comprising $n$ tokens and $m$ sub-structure sequences. Then, $y$ can be represented at the token level as $y = t_1, t_2, ... t_n$, or at the sub-structure level as $y = s_1, s_2, ... s_m$.

During LLM generation, the sub-structure sequence can be split by pre-defined special tokens based on the format of the generated structured data. For instance, in product attribute generation, two special tokens ``<SEP>'' and ```<END>'', can be defined to separate the attribute key and value and different attributes, respectively. An attribute $s$ then can be represented by the following sequence of tokens:

\begin{equation}
s = t_{k, 1}, t_{k, 2}, ... \text{<SEP>}, t_{v, 1}, t_{v, 2}, ... \text{<END>}
\end{equation}

where $t_{k, i}$ and $t_{v, j}$ are the $i$-th token in the attribute key and $j$-th token in the attribute value, respectively. 

\subsection{Confidence Estimation in Structured Data Generation}

\begin{figure}[ht]
    \includegraphics[width=\columnwidth]{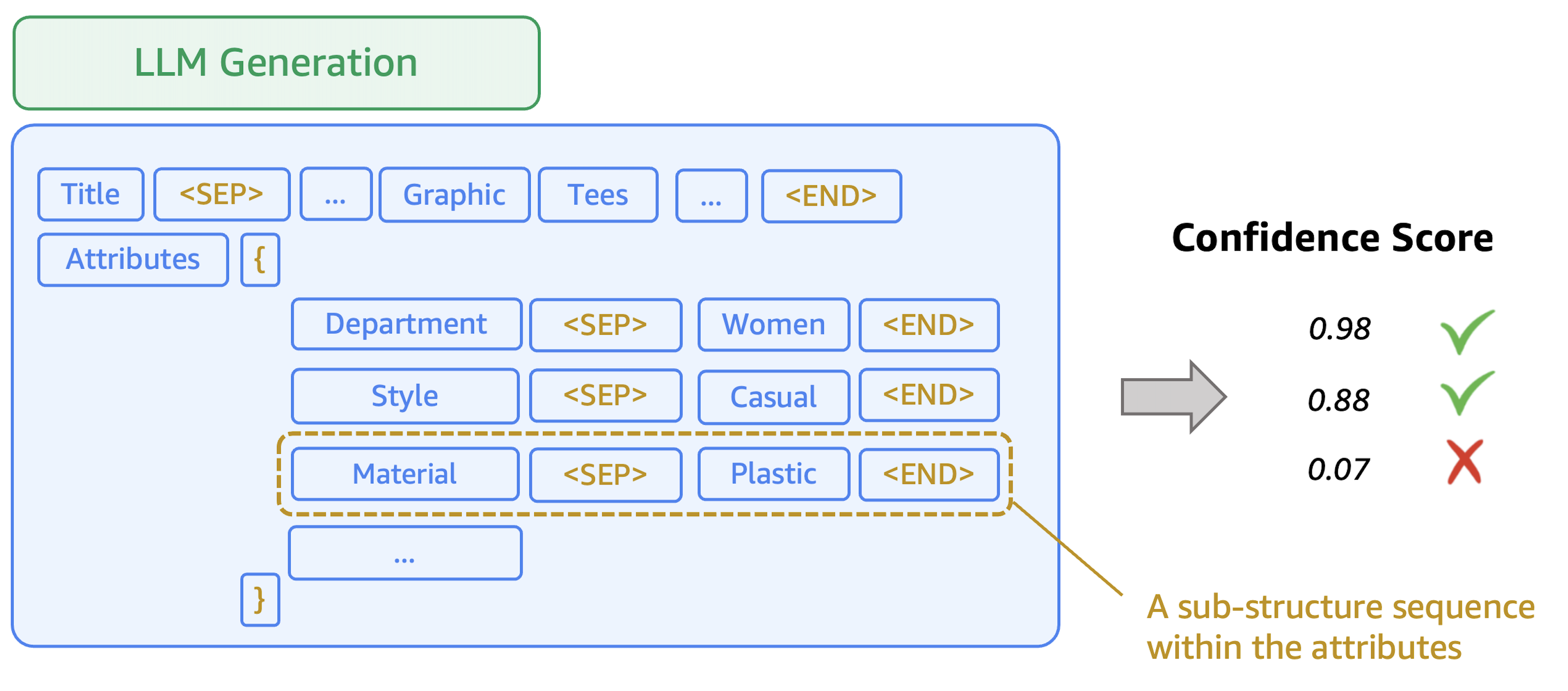}
    \caption{In structured data generation, we tokenize the LLM output into sub-structure sequences and assign a confidence score to each prediction based on the prescribed method.}
    \label{fig:conf_pred_in_structured_data}
\end{figure}

In the structured data generation process, LLMs may produce incorrect predictions in certain sub-structure-level data due to inaccurate inferences or hallucinations. As the example of product attribute generation illustrated in Fig.~\ref{fig:conf_pred_in_structured_data}, the sub-structure sequence ``Material <SEP> Plastic <END>'' is incorrectly generated while the remaining sub-structure sequences are correct (e.g., ``Department <SEP> Women <END>'', ``Style <SEP> Casual <END>''). Employing a confidence estimator becomes crucial to address such instances. This estimator assigns a score to each sub-structure sequence, enabling the identification of incorrect predictions and enhancing the overall reliability of the generated structured data.

%%%%%%%%%%%%%%%%%%%%%%%%%%%%%%%%%%%%%%%%%%%%%%%%%%%%%%%%%%%%%%%%%%
%%%%%%%%%%%%%%%%%%%%%%%%%%%%%%%%%%%%%%%%%%%%%%%%%%%%%%%%%%%%%%%%%%

\subsection{Confidence Estimation Method} \label{subsec:confidence methods}

Two categories of methods for obtaining confidence scores for sub-structure sequences are studied. The first utilizes token conditional probability, while the second trains a confidence estimator based on the internal hidden states from LLMs.

\subsubsection{Token Conditional Probability}

The token's conditional probability can be determined through the logits in the final layer outputted by the auto-regressive LLM. The conditional probability $p(t_i|t_{<i}, x)$ of token $i$ is calculated using the Softmax function.
% \begin{equation}
% p(t_i|x) = \frac{\exp(\text{logit}_i)}{\sum_{j=1}^{V} \exp(\text{logit}_j)}
% \end{equation}
The token conditional probability can be the source to measure how confident the model generates a sub-structure sequence. We define the confidence method \textbf{CP}, which computes the confidence score for sub-structure sequence $\text{s}$ as the product of the conditional probabilities of its tokens:

\begin{equation}
\text{conf}(\text{s}) = \prod_{i=1}^{n} p(t_i|t_{<i},x), \quad t_i \in \text{s} 
\end{equation}

Additionally, we consider length normalization for the CP method, which is normalized by the length of the sub-structure sequence. This method is denoted as \textbf{CP w/ LN} and is calculated by:

\begin{equation}
\text{conf}(\text{s}) = \left(\prod_{i=1}^{n} p(t_i|t_{<i},x)\right)^{1/n}, \quad t_i \in \text{s}
\end{equation}

% One could consider doing normalization to the conditional probability, for instance, through temperature scaling. However, these approaches essentially do not alter the monotonic relationship of the confidence score. Consequently, they do not yield significant differences in terms of the average precision and recall at certain precision, which are important evaluation metrics we used in this work. As a result, we have chosen not to include them in this study.

% The temperature-based scaling method, denoted as \textbf{Temp.} is another confidence method based on the token probability. The only difference from \textbf{Cond. Prob.} is that the token conditional probability is scaled by a temperature parameter $T$.

% \begin{equation}
%     p(t_i|x) = \frac{\exp(\text{logit}_i/T)}{\sum_{j=1}^{V}\exp(\text{logit}_j/T)}, \quad T > 0
% \end{equation}

\subsubsection{Confidence Network}

We introduce the Confidence Network, a Feed Forward Network classifier, as the confidence estimator for sub-structure sequences using the internal hidden states of an LLM. 
The Confidence Network is trained as a binary classifier to predict whether if the generated sub-structure sequence is faithful. Specifically, the hidden states from an LLM are utilized to construct the representation of the sub-structure sequence, which serves as the input for the Confidence Network. During inference, the soft label outputted by the Confidence Network is then used as the confidence score for the sub-structure sequence generated by the LLM.

As hidden states from different layers of an LLM contain different perspectives of information, hidden states from various layers are tested. Given a sub-structure sequence $s$ consisting of $p$ tokens, $\{t_1, t_2, \ldots, t_p\}$, we can retrieve the corresponding hidden state of layer $l$ $\{h_{1l}, h_{2l}, \ldots, h_{pl}\}$ from the LLM. To derive the sub-structure representation from these internal hidden states, we employ three different methods following \cite{toshniwal2020cross}:

\begin{itemize}
\setlength{\itemsep}{0pt}
\setlength{\parskip}{0pt}
  \item \textbf{Last} corresponds to the hidden state of the last token in the sub-structure sequence, i.e., $[h_{pl}]$
  \item \textbf{Extreme} involves the concatenation of the hidden states of the first and last tokens in the sub-structure sequence, i.e., $[h_{1l}; h_{pl}]$
  \item \textbf{Sum-Diff} concatenates the sum and difference of the hidden states of the first and last tokens in the sub-structure sequence, i.e., $[h_{1l}+h_{pl}; h_{1l}-h_{pl}]$
\end{itemize}

%%%%%%%%%%%%%%%%%%%%%%%%%%%%%%%%%%%%%%%%%%%%%%%%%%%%%%%%%%%%%%%%%%
\begin{figure*}[ht]
\begin{center}
\includegraphics[width=0.95\textwidth]{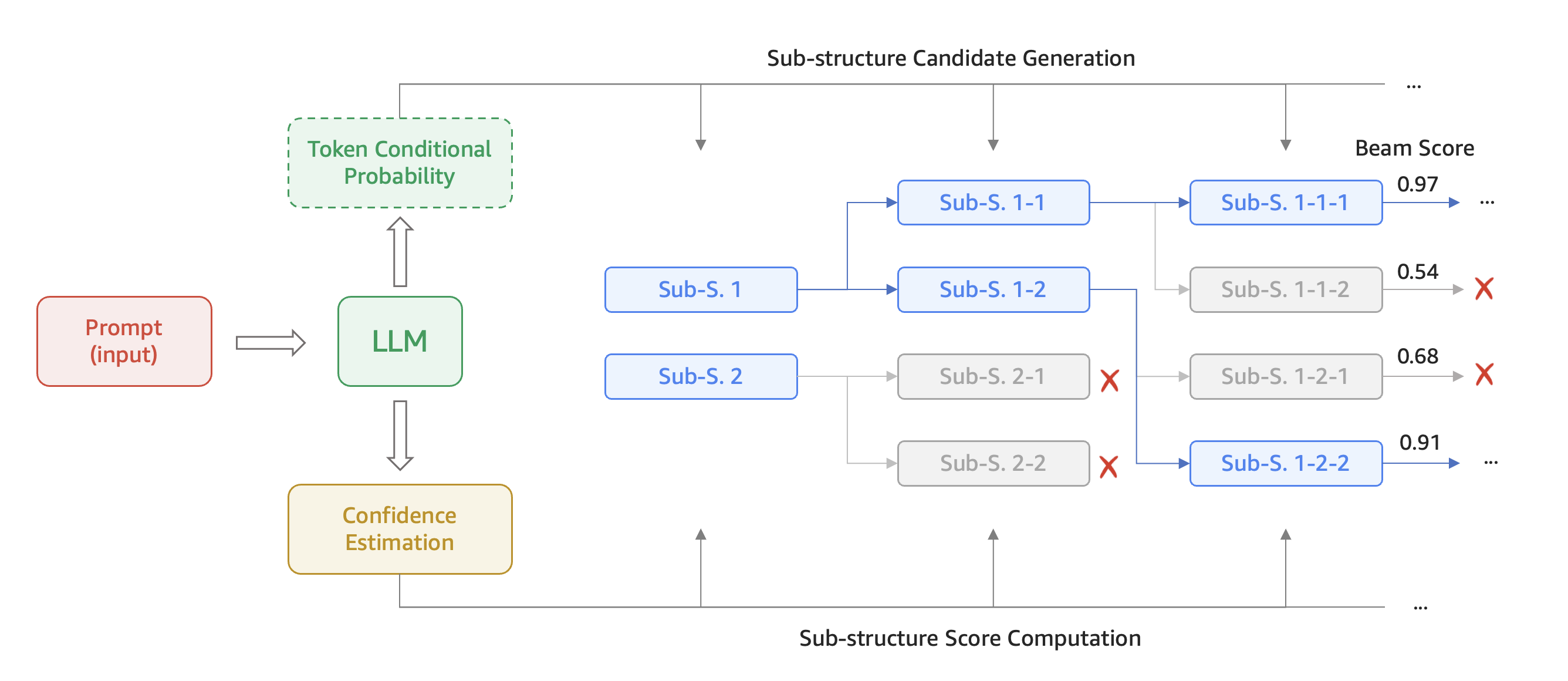}
\caption{Confidence-aware sub-structure Beam Search. The beam size is set to 2 for illustration. In the decoding process, beams with higher scores (highlighted in blue) are kept.}
\label{fig:cabs}
\end{center}
\end{figure*}
%%%%%%%%%%%%%%%%%%%%%%%%%%%%%%%%%%%%%%%%%%%%%%%%%%%%%%%%%%%%%%%%%%

\subsection{Confidence-Aware Sub-Structure Decoding}

We further propose Confidence-Aware sub-structure Beam Search (\textbf{CABS}). CABS is suitable for the generation of structure data as it operates on the sub-structure level and considers the confidence score for each sub-structure sequence.

Greedy search is one of the most common decoding methods in text generation. As illustrated in Equation~\ref{eq: greedy search}, when generating a sequence of structured data $y$ with $n$ tokens, it selects the token with the maximum conditional probability at each time $i$.

\begin{equation} \label{eq: greedy search}
    \prod_{i=1}^{n}\arg \max_{t_i} p(t_i|t_{<i},x), t_i \in V
\end{equation}

where $V$ is the vocabulary.

Traditional beam search operates on the token level and picks the path with the highest probability among the hypotheses of beam size:

\begin{equation}
    \arg \max_{y} \prod_{i=1}^{n} p(t_i|t_{<i},x), t_i \in V
\end{equation}

However, in the context of structured data generation, solely considering token-level information during generation disregards the structural format information inherent in structured data. There is a desire to operate not only at the token level but also at the sub-structure level. Additionally, identifying and excluding unfaithful sub-structure generation during the decoding process can prevent subsequent sub-structure generation from being impacted by previous incorrect generation. Thus, we adapt traditional token-level beam search into CABS, with the aim of evaluating the confidence scores of sub-structure sequences and generating structured data with the maximum total confidence score across all sub-structure sequences. Assuming the structured data $y$ has $m$ sub-structure sequences, one can simply compute the confidence score of a sub-structure sequence using the CP confidence estimation method, i.e., taking the product of token conditional probabilities, as illustrated in Equation~\ref{eqn:attr_level_beam_search}:

\begin{equation} \label{eqn:attr_level_beam_search}
    \arg \max_{y} \prod_{j=1}^{m} conf(s_j), s \in \text{SUB-S} 
\end{equation}
\[
conf(s_j) = \prod_{i=1}^{n} p(t_i|t_{<i},x), t_i \in s_j
\]

where $SUB-S$ is the set of sub-structures candidates, and it can be generated by token-level beam search. 

% The sub-structure beam search can be seen as utilizing the \textbf{CP} confidence score to select the next attribute for generation. We denote this decoding method as \textbf{CABS-CP}. 
Since purely depending on the token conditional probability might not solve the unfaithful sub-structure generation efficiently, we can further use the confidence score predicted by the Confidence Network in CABS. Figure~\ref{fig:cabs} illustrates the workflow of CABS. At each step in generating the next sub-structure sequence, CABS first generates sub-structure candidates using token-level beam search. It then utilizes the confidence score predicted by a confidence estimation method, the CP or Confidence Network, as the score for each sub-structure, and retains the path with a higher score. 

% This results in the formulation of CABS, as shown in Equation~\ref{eq: CABS}.
% \begin{equation} \label{eq: CABS}
%     score(s_j) = \prod_{i=1}^{n} p(t_i|t_{<i},x) + \beta * conf(s_j), \; t_i \in s_j 
% \end{equation}

\section{Model \& Data} \label{sec: data and model}

We validate our approach in the domain of structured product catalog data. We formulate a generation problem where the LLM is tasked with generating complete product attributes given low-quality product entity data, such as incorrect or missing attribute values. This scenario is similar to the example illustrated in Fig.~\ref{fig:example_structured_data_generation} (b), but within the context of product catalogs.

\subsection{Experimentation Setup}

%%%%%%%%%%%%%%%%%%%%%%%%%%%%%%%%%%%%%%%%%%%%%%%%%%%%%%%%%%%%%%%%%%
\begin{figure*}[t]
    \includegraphics[width=0.75\textwidth]{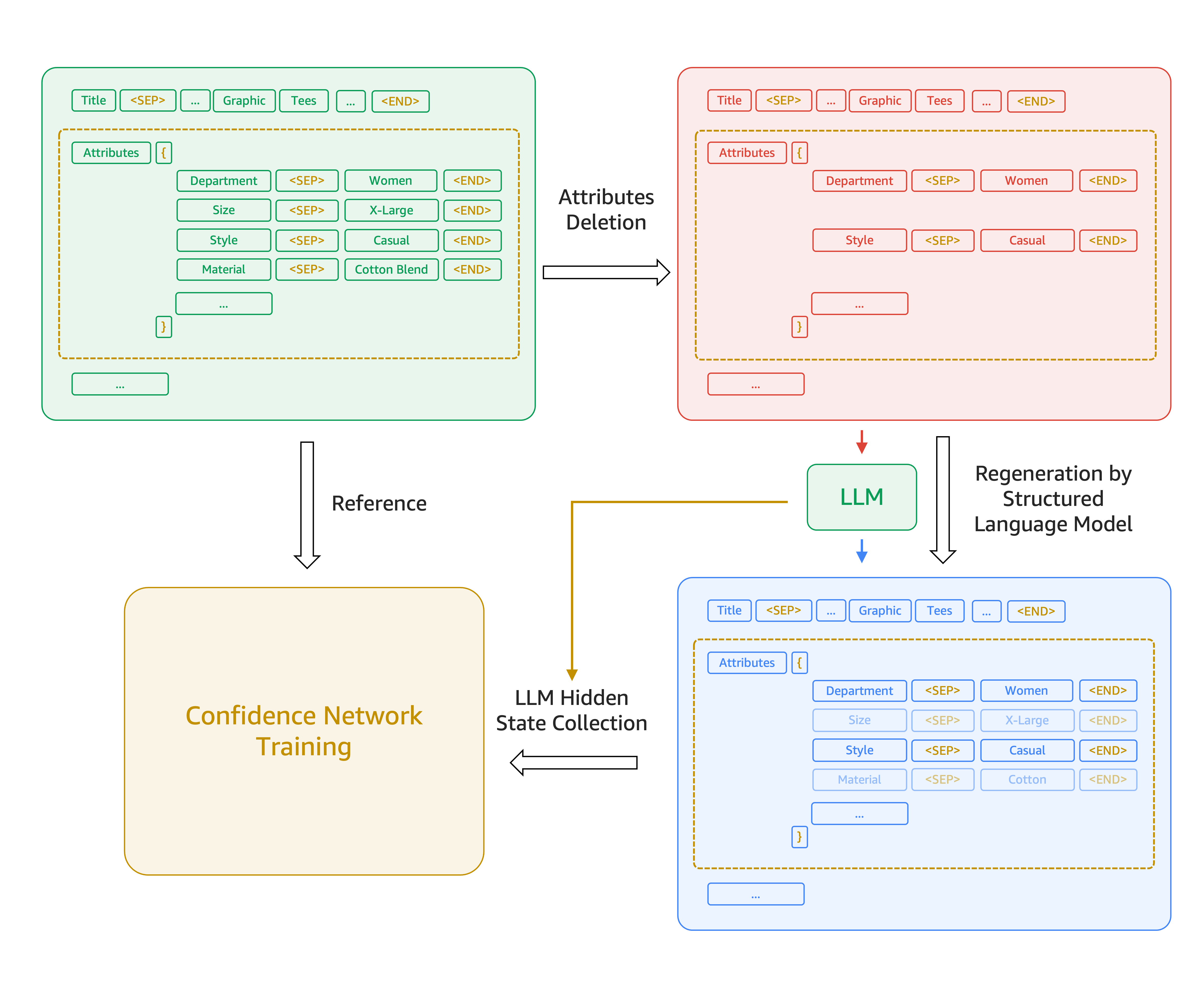}
    \caption{Attribute deletion for self-supervised training the Confidence Network. Light blue denotes the incremental attributes generated by the LLM. We use the original attribute values as a reference to determine if the LLM generated values are correct.}
    \label{fig:conf-predictor self-supervised-training}
\end{figure*}
%%%%%%%%%%%%%%%%%%%%%%%%%%%%%%%%%%%%%%%%%%%%%%%%%%%%%%%%%%%%%%%%%%

\textbf{Backbone Structured Data Generation LLM}. We use a fine-tuned Product Catalog LLM based on the publicly available MPT 7B~\cite{MosaicML2023Introducing} as the backbone LLM. The fine-tuned Product Catalog LLM is able to generate all relevant product attributes in a single pass, given product entity data. 

\textbf{Confidence Network Model Setup and Training Strategy.}
In this paper, we used a 3-layer Feed Forward Network classifier as the Confidence Network. The Confidence Network applies ReLU non-linear activation between hidden layers and Sigmoid non-linear activation after the last layer. It is trained using the binary cross-entropy loss function, aiming to predict the faithfulness of the generated sub-structure sequence.

Training the Confidence Network requires labels for each generated sub-structure-level data, i.e., for each product attribute value generated by the LLM, we need a label to indicate whether the generated attribute is correct. Manually labeling this data set at scale is both costly and time-consuming. Instead, we adopt a data corruption strategy or, more specifically, attribute deletion to train the sub-structure confidence predictor in a self-supervised manner as shown in Fig. \ref{fig:conf-predictor self-supervised-training}.
At a high level, our goal is to delete known attributes which are audited as correct and allow the LLM to regenerate the attributes on its own. For the purpose of training the confidence network, 20\% of known and correct attribute values are randomly deleted from the prompt. 
Based on the modified prompt, the fine-tuned Product Catalog LLM is tasked with re-generating all relevant product attributes. The originally known attribute values are then employed as a reference to determine the correctness of the generated attributes. In the examples shown in Fig. \ref{fig:conf-predictor self-supervised-training}, ``Size <SEP> X-Large <END>'' generated by the LLM is considered as correct as it matches with the original attribute value, while ``Material <SEP> Cotton <END>'' does not.

%, using exact string match. Lastly, the attribute confidence predictor is trained using the attribute representation formed by hidden state to determine the generated attribute is correct or not. 

\subsection{Data}

\textbf{Confidence Network Training Data}: The confidence network is trained with 1.5 million attributes from 400k English products sampled from an established e-commerce store. Examples of product attributes include Brand, Color and Size. In addition, product attributes must conform to a predefined schema. For example, the attribute ''Assembly Required'' only accepts true/false values.

\textbf{Test Data}. We randomly sampled a test set containing 1k English products with 10k product attributes. The LLM generated attribute values are labeled by a group of in-house auditors who are trained for this domain.

\section{Experimental Results \& Analysis} \label{sec: experiments}

\subsection{Evaluation Metric}
Average precision serves as a metric for evaluating performance across various sub-structure confidence estimation and decoding methods. Furthermore, we assess recall at a specific precision since many real-world applications aim to guarantee that the generated sub-structure-level data is above a designated precision threshold while maintaining a high level of recall.

%%%%%%%%%%%%%%%%%%%%%%%%%%%%%%%%%%%%%%%%%%%%%%%%%%%%%%%%%%%%%%%%%%
\begin{figure}[b]
    \includegraphics[width=0.95\columnwidth]{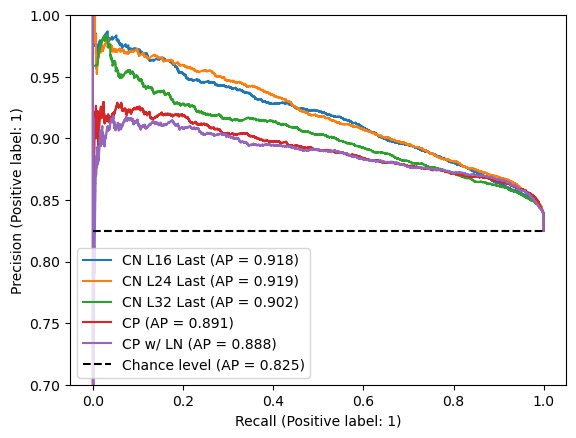}
    \caption{PR curve and Average Precision of confidence methods. Confidence Network (CN) methods that only use 'Last' attribute representation are displayed to enhance visualization}
    \label{fig:conf-predictor results}
\end{figure}
%%%%%%%%%%%%%%%%%%%%%%%%%%%%%%%%%%%%%%%%%%%%%%%%%%%%%%%%%%%%%%%%%%

\subsection{Confidence Estimation Method Results}

The performance of confidence estimation methods in Sec.~\ref{subsec:confidence methods} is first evaluated in this subsection.

\textbf{Comparison between the token conditional probability and Confidence Network}.
The PR curve and average precision for token conditional probability and Confidence Network confidence estimation methods are shown in Fig.~\ref{fig:conf-predictor results}. The Confidence Network methods using the \textbf{Last} sub-structure representation from hidden layers 16, 24, and 32 in the 32-layer MPT-7b model are displayed. We observe that the Confidence Network methods significantly outperform the token conditional probability ones, as enhanced by the self-supervised training. Furthermore, the PR curves suggest that, as we adjust the threshold, the LLM's generation is capable of achieving higher precision with a certain sacrifice on recall. 

\textbf{Effect of different layers and sub-structure representations}. Table~\ref{table: sub-structure performance} displays the average precision and recall at 90\% precision for sub-structure representations constructed from the hidden states of layers 16, 24, and 32. First, sub-structure representations built using Sum-Diff underperform when compared to Last and Extreme representations. Furthermore, we found out that hidden states from middle layer outputs outperform those from the last layer. This is consistent with the finding in~\cite{azaria2023internal} where the last layer does not perform the best. In our case, we conjecture that this is due to the data difference between training and testing. We introduce synthetic noise, namely attribute deletion, into the training data, while the noise in the test data is more closely aligned with the real world. Hidden states from Layer 32 are prone to overfitting the training data for the Confidence Network.

% %%%%%%%%%%%%%%%%%%%%%%%%%%%%%%%%%%%%%%%%%%%%%%%%%%%%%%%%%%%%%%%%%%
% \begin{table}[ht]
%   \caption{Average precision and R@P90 of Confidence Network using different layers and sub-structure representations. The best results are shown in bold, while the second best results are underlined.}
%   \label{table: average precision}
%   \begin{tabular}{ccccccc}
%     \toprule
%     \multirow{2}{*}{\shortstack[c]{Sub-Structure \\ Representation}} & \multicolumn{2}{c}{L16} & \multicolumn{2}{c}{L24} & \multicolumn{2}{c}{L32} \\ \cline{2-7}
%       & AP & R@P90 & AP & R@P90 & AP & R@P90 \\
%     \midrule
%     Last & \underline{91.8} & 64.5 & \textbf{91.9} & 65.8 & 90.2 & 53.9 \\
%     Extreme & \underline{91.8} & \underline{64.7} & \underline{91.8} & \textbf{67.9} & 90.7 & 56.8\\
%     Sum-Diff & 91.1 & 58.7 & 90.8 & 59.2 & 90.3 & 50.4 \\
%   \bottomrule
% \end{tabular}
% \end{table}
% %%%%%%%%%%%%%%%%%%%%%%%%%%%%%%%%%%%%%%%%%%%%%%%%%%%%%%%%%%%%%%%%%%

%%%%%%%%%%%%%%%%%%%%%%%%%%%%%%%%%%%%%%%%%%%%%%%%%%%%%%%%%%%%%%%%%%
\begin{table}[ht]
  \caption{Average precision and R@P90 of Confidence Network using different layers and sub-structure representations. The best results are shown in bold, while the second-best results are underlined.}
  \label{table: sub-structure performance}
  \begin{tabular}{cccc}
    \toprule
    AP / R@P90   & L16 & L24 & L32 \\
    \midrule
    Last & \underline{91.8}/64.5 & \textbf{91.9}/65.8 & 90.2/53.9 \\
    Extreme & \underline{91.8}/\underline{64.7} & \underline{91.8}/\textbf{67.9} & 90.7/56.8\\
    Sum-Diff & 91.1/58.7 & 90.8/59.2 & 90.3/50.4 \\
  \bottomrule
\end{tabular}
\end{table}
%%%%%%%%%%%%%%%%%%%%%%%%%%%%%%%%%%%%%%%%%%%%%%%%%%%%%%%%%%%%%%%%%%

\subsection{CABS Results}

CABS can employ either CP or the Confidence Network to obtain confidence scores for sub-structure sequences. To ensure an unbiased evaluation of each generated sub-structure sequence, we utilize a different confidence estimation method to assess the confidence score of CABS's generation than the one employed in CABS. Specifically, if CABS utilizes confidence scores from the Confidence Network during decoding, we evaluate its generation performance using CP, and vice versa. The comparison between CABS and other decoding methods is presented in Table~\ref{tab: decoding results}. In our experiments, the Confidence Network applies Extreme attribute representation built on the hidden state of layer 16.

Firstly, experimental results show that CABS significantly outperforms greedy search and token-level beam search decoding methods in terms of both average precision and recall at 90\% precision. That performance improvement is attributed to CABS operating at the sub-structure level and assessing the confidence score for each sub-structure during generation. Secondly, we observed that a larger beam size can enhance the performance of both token-level beam search and CABS due to a larger search space.

% Implicit / Explicit Attributes
% PROB = (conditional probability/confidence network)

%%%%%%%%%%%%%%%%%%%%%%%%%%%%%%%%%%%%%%%%%%%%%%%%%%%%%%%%%%%%%%%%%%
\begin{table}
\caption{Average Precision and R@P90 of decoding methods. The confidence scores obtained by token conditional probability and Confidence Network are both included.}
\label{tab: decoding results}
\centering
\begin{tabular}{cccccc}
\toprule
\multirow{3}{*}{\shortstack[c]{Decoding \\ Method}} & \multirow{3}{*}{Beam Size} & \multicolumn{4}{c}{Confidence Estimation} \\ \cmidrule{3-6}
 & & \multicolumn{2}{c}{CP} & \multicolumn{2}{c}{CN} \\ \cmidrule(lr){3-4} \cmidrule(lr){5-6}
 & & AP & R@P90 & AP & R@P90 \\ 
\midrule
Greedy Search & - & 89.1 & 36.6 & 91.8 & 64.7 \\
\midrule
\multirow{2}{*}{\vspace{-0.5em} \shortstack[c]{Token-Level \\ Beam Search}} & 2 & 89.3 & 34.1 & \underline{92.1} & 66.8 \\
 & 4 & 89.3 & 34.8 & \underline{92.1} & 67.0 \\
\midrule
\multirow{2}{*}{CABS} & 2 & \textbf{90.8} & \underline{58.6} & \textbf{92.6} & \underline{74.7} \\
 & 4 & \underline{90.5} & \textbf{60.5} & \textbf{92.6} & \textbf{75.6} \\
\bottomrule
\end{tabular}
\end{table}
%%%%%%%%%%%%%%%%%%%%%%%%%%%%%%%%%%%%%%%%%%%%%%%%%%%%%%%%%%%%%%%%%%

% %%%%%%%%%%%%%%%%%%%%%%%%%%%%%%%%%%%%%%%%%%%%%%%%%%%%%%%%%%%%%%%%%%
% \begin{table}
% \caption{PR curve and Average Precision of decoding methods. The confidence score obtain by token conditional probability and Confidence Network are both included.}
% \label{tab: decoding results}
% \centering
% \begin{tabular}{cccc}
% \hline
% \multirow{2}{*}{Decoding Method} & \multirow{2}{*}{Beam Size} & \multicolumn{2}{c}{Confidence Estimation} \\\cline{3-4}
%  & & CP & CN \\
% \hline
% Greedy Search & - & 89.1/36.6 & 91.8/64.7 \\
% \hline
% \multirow{2}{*}{\vspace{-0.5em} \shortstack[c]{Token-Level \\ Beam Search}} & 2 & 89.3/34.1 & 92.1/66.8 \\
%  & 4 & 89.3/34.8 & 92.1/67.0 \\
% \hline
% \multirow{2}{*}{CABS} & 2 & 90.8/58.6 & \textbf{92.6}/74.7 \\
%  & 4 & 90.5/60.5 & \textbf{92.6}/\textbf{75.6} \\
% \hline
% \end{tabular}
% \end{table}
% %%%%%%%%%%%%%%%%%%%%%%%%%%%%%%%%%%%%%%%%%%%%%%%%%%%%%%%%%%%%%%%%%%

\section{Related Work} \label{sec: related work}

\subsection{Model Confidence Estimation}

Model confidence estimation methods measure the probability of the model's prediction being correct, reflecting the trustworthiness of the model's output. Existing work mainly works on complete sequence-level or token-level confidence estimation. 
For complete sequence-level confidence estimation, \citet{jiang2021can} studied uncertainty estimation on the task of question answering, employing methods including fine-tuning, post-hoc probability modification, and adjustment of the predicted outputs or inputs. \citet{azaria2023internal, kadavath2022language} trained an additional model to predict the LLM's statement truthfulness based on the hidden layer activation. \citet{malinin2020uncertainty} examined the uncertainty estimation on machine translation task. \citet{lin2022teaching, manakul2023selfcheckgpt} treated the language model as a black box, designing specific tasks or datasets to obtain model confidence based on the text output without utilizing logits from the model.
For token-level confidence estimation, \citet{xiao2021hallucination, petryk2024simple} made a token-level confidence estimation case study on image captioning.  \citet{zhou2021detecting} explored the algebraic and learning methods for token-level confidence in the context of machine translation and abstractive summarization. 
Neither sequence-level nor token-level confidence estimation methods are suitable for structured data generation, as they lack sub-structured-level confidence to distinguish between generated sub-structures where the LLM is confident and where it is not.

% Model confidence, the probability of the model's prediction being correct, reflect the trustworthiness of the model's output. On the entire sequence level, . The most direct method to obtain language model uncertainty is to use the logits output from the model. \citet{jiang2021can} studied uncertainty estimation of the sequence generated by LLM on the task of question answering, including fine-tuning, post-hoc probability modification, and adjustment of the predicted outputs or inputs. \citet{xiao2021hallucination} made a token-level confidence estimation case study on image captioning. They calculate token uncertainty by the entropy of the token probability distributions a model predicts, and consequently proposed a uncertainty-aware beam search algorithm to reduce hallucination. \citep{azaria2023internal, kadavath2022language} trained an additional model to predict the LLM's statement truthfulness based on the hidden layer activations. Other methods \cite{lin2022teaching, manakul2023selfcheckgpt} treat LM as a black box and design specific task or dataset to obtain the model confidence based on the text output of model without using the logit from the model. 

\subsection{Text Generation Decoding Methods}. 
\vspace{-1ex}

Most text generation decoding methods work on the token probability. That is, they generate text solely based on token probability. In deterministic decoding methods, greedy search chooses the token with the highest probability. Beam search is introduced to choose a text sequence with the overall highest probability instead of choosing the token with the highest probability at each step. In stochastic decoding methods, Top-K \cite{fan2018hierarchical}, which samples the next generated token within the tokens with top K probabilities, and Top-P sampling \cite{holtzman2019curious}, which samples from the smallest token set whose cumulative probability exceeds a certain threshold probability, are proposed to increase the diversity of generated text. 

On the other hand, many NLP applications require other control signals during the text generation, instead of solely based on token probability. \citet{shen2022sentbs} proposed a sentence-level beam search method to control sentence generation with respect to the text structure. \citet{chen2023sentence} introduced a tree search method that operates on sentence level for text generation. \citet{xiao2021hallucination} proposed an uncertainty-aware beam search algorithm to reduce LLMs' hallucination, which shares some similarities with our work. Nevertheless, our work distinguishes itself as operating on a sub-structure level, focusing on structured data generation.

\section{Conclusion} \label{sec: conclusion}

This work investigates the confidence estimation methods for LLMs in generating structured data. Through experiments, we find LLM internal states effective in assessing sub-structure data faithfulness. Furthermore, we introduce a novel decoding method called CABS. It enhances structured data generation faithfulness by assessing the confidence score for each generated sub-structure-level data and iteratively refining prompts. As to future work direction, it's promising to extend the application of CABS to more types of structured data, such as tabular data and knowledge bases. Additionally, exploring a more advanced confidence estimation method to further boost the performance of CABS is a promising research avenue.

%%
%% The next two lines define the bibliography style to be used, and
%% the bibliography file.
%\newpage
\bibliographystyle{ACM-Reference-Format}
\bibliography{ref}

%%
%% If your work has an appendix, this is the place to put it.
\appendix

\end{document}